\documentclass{article}
\usepackage{spconf,amsmath,graphicx,hyperref}
\usepackage{algorithm}
\usepackage{algorithmic}
\usepackage{amsfonts}
\usepackage{booktabs}
\usepackage{multirow}
\usepackage{array} 
\usepackage{dashrule} 
\usepackage{graphicx}
\usepackage{xcolor}
\usepackage{enumitem}
\usepackage{threeparttable}

\renewcommand{\arraystretch}{1}   % 行高

\newcommand{\equalcontrib}{\textsuperscript{*}}
\newcommand{\one}{\textsuperscript{1}}
\newcommand{\two}{\textsuperscript{2}}

\title{DSFT: Inspiring Diffusion Large Language Models to Comprehend Mathematical and Logical Patterns}
\name{Ranfei Chen\one, Ming Chen\two\equalcontrib \thanks{* Corresponding author.}}
\address{\one Key Laboratory of Intelligent Information Processing, \\ Institute of Computing Technology, Chinese Academy of Sciences, Beijing, China \\ 
\two Computer Network Information Center, Chinese Academy of Sciences, Beijing, China}
\begin{document}
\ninept
\maketitle
\begin{abstract}
Diffusion large language models (dLLMs) have emerged as a new architecture following auto regressive models. Their denoising process offers a powerful generative advantage, but they present significant challenges in learning and understanding numerically sensitive mathematical and order-sensitive logical tasks. Current training methods, including pre-training, fine-tuning, and reinforcement learning, focus primarily on improving general knowledge retention and reasoning abilities, but lack a comprehensive understanding of mathematical and logical patterns. We propose DSFT, a simple yet effective Diffusion SFT strategy, by adjusting the masking strategy and loss function, guiding models to understand mathematical and logical patterns. This strategy can be flexibly combined with pre-training, reinforcement learning, and other training methods. Validated on models such as LLaDA and Dream series, we prove that DSFT on small-scale data can achieve improvements of 5-10\% and approximately 2\% on mathematical and logical problems, respectively. This inspiring masking approach offers insights for future learning of specific patterns, which can be easily and efficiently combined with other training methods and applied to various dLLMs. Our code is publicly available at \url{https://anonymous.4open.science/r/DSFT-0FFB/}.
\end{abstract}

\section{Introduction}
\label{sec:intro}

The paradigm of large language models has been largely defined by the sequential, token-by-token generation of auto regressive architectures. However, a new class of diffusion Large Language Models (dLLMs) has recently emerged as a viable and powerful alternative. This exciting field has seen rapid progress, dLLMs beginning with pioneering works that ingeniously adapted diffusion principles from continuous domains to the discrete nature of text, establishing fundamental approaches through both continuous embeddings \cite{li2022diffusion} and discrete state-space transitions \cite{austin2021structured}. 

Fueled by this initial spark, the research community has achieved remarkable strides in scaling and refining dLLMs, demonstrating their potential to rival or even surpass AR-LLM in diverse linguistic tasks. Landmark models such as LLaDA \cite{nie2025large}—which pioneered large-scale diffusion through masked denoising processes and its successor LLaDA-1.5 \cite{zhu2025llada}, which incorporated variance-reduced preference optimization for enhanced training stability and generation quality, have successfully expanded dLLMs to billion-parameter regimes. Complementing these are innovative frameworks like Dream-7B \cite{ye2025dream}, a powerful diffusion model that matched the performance of strong autoregressive models on general tasks while demonstrating superior capabilities in complex planning and reasoning, and DiffuCoder \cite{gong2025diffucoder}, which uses diffusion techniques to unlock new potentials in code synthesis and programming applications. This progress is supported by a robust wave of theoretical advancements, including the establishment of scaling laws for masked diffusion models in SMDM \cite{nie2025scaling}, simplified yet effective architectures in MDLM \cite{sahoo2024simple} and EDLM \cite{xu2025energybased}, and time-independent parameterizations via concrete scores in RADD \cite{ou2024your}, alongside deeper insights into discrete diffusion dynamics \cite{lou2024discrete}. This theoretical foundation is further strengthened by a vibrant open-source ecosystem, exemplified by Open-dLLM \cite{opendllm2025}, which provides comprehensive tools for training, evaluation, and deployment, fostering widespread adoption and collaborative innovation. Meanwhile, practical deployments are being accelerated through specialized techniques, such as the Fast-dLLM framework \cite{wu2025fast} from NVIDIA, which uses KV caching, parallel decoding, and other optimizations to boost inference speeds, alongside enhancements in sampling strategies like path planning \cite{peng2025path} and remasking for iterative token refinement, and masking improvements via path pruning for greater efficiency and coherence.

\begin{figure}[t]
    \centering
    \includegraphics[width=0.48\textwidth]{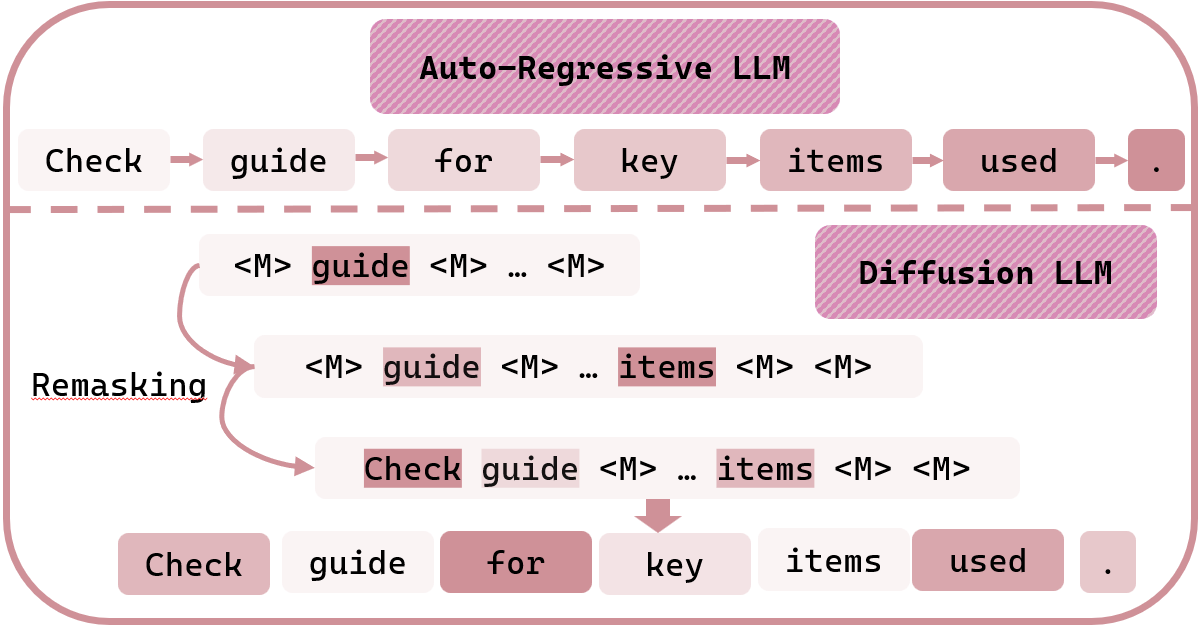}
    \vspace{-2em}
    \caption{Auto Regressive LLM and Diffusion LLM}
    \label{fig:framework_overview}
\end{figure}

\begin{figure*}[t]
    \centering
    \includegraphics[width=0.8\textwidth]{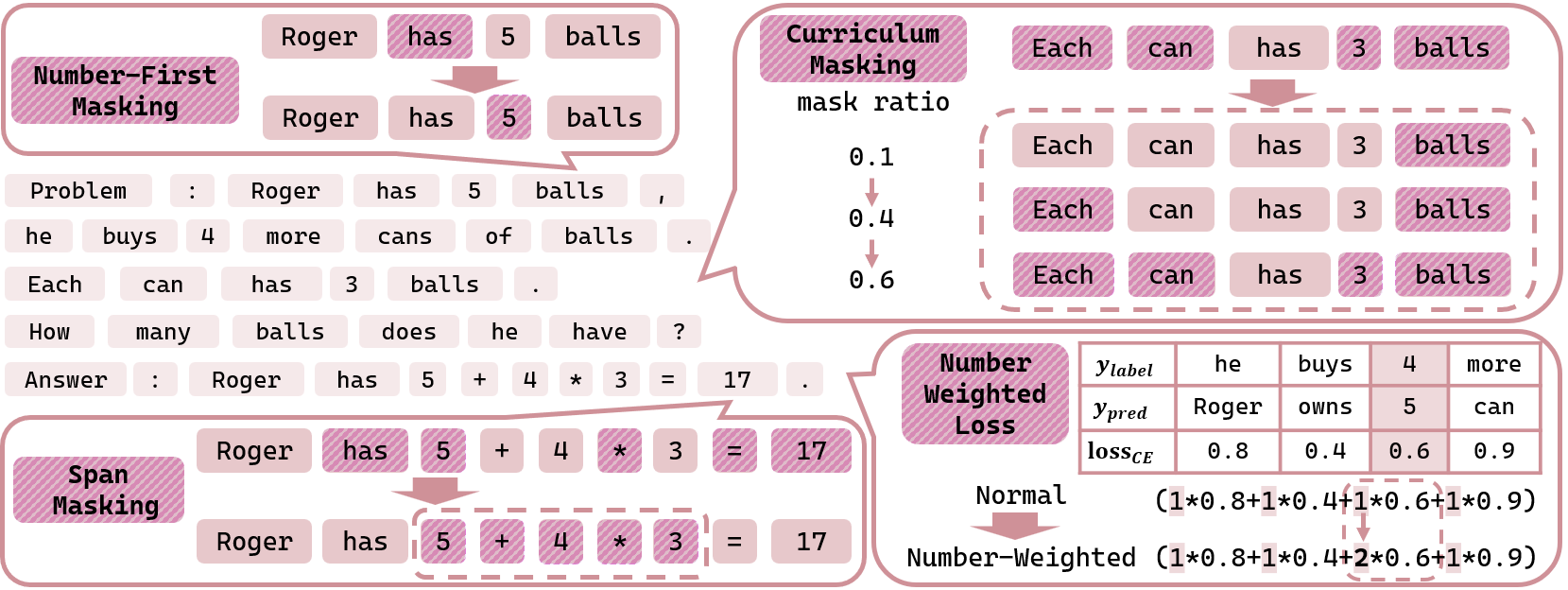}
    \caption{An example showing how the DSFT method works with its four fine-tuning techniques illustrated: (1) \textbf
    {Number-First Masking} focuses on key numerical 
    values; (2) \textbf{Span Masking} targets 
    contiguous expressions like the core calculation; (3) 
    \textbf{Curriculum Masking} shows an increasing mask 
    ratio over time; (4) \textbf{Number-Weighted Loss} 
    applies higher weights to numerical tokens during 
    training.}
    \label{fig:methodology_overview}
\end{figure*}

With dLLMs now established as powerful general-purpose models, the next critical challenge is to release their potential in specialized tasks. The standard approach, Supervised Fine-Tuning. However presents a fundamental limitation. SFT is often a semantically-blind process, applying a uniform training signal that treats all tokens—from the number '7' to the word 'the' with equal importance. This approach fails to capture the unique grammar and semantic hierarchy of specialized domains such as mathematics, leading to superficial pattern-matching rather than deep understanding. The critical question, therefore, is how we can move beyond uniform fine-tuning to inspire a genuine understanding of the unique patterns of a specialized domain.

So we introduce \textbf{DSFT (Diffusion Supervised Fine-Tuning)}, a methodology that acts as a cognitive lens, guiding model's attention toward what truly matters in mathematical text. Our approach is grounded in a simple information-theoretic principle: in mathematical text, tokens like numbers and operators have higher information entropy, which are more critical for comprehension. DSFT implements this insight through four simple yet effective techniques: (1) Number-First Masking, (2) Span Masking, (3) Curriculum Difficulty Progression, and (4) Number Weighted Loss. 

Collectively, these techniques guide the model towards different learning approaches based on its sensitivity to specific information. We validate this methodology through extensive experiments and prove that DSFT significantly improves mathematical and logical comprehension on top of leading dLLMs using only a small dataset. Ablation studies confirm the synergistic contribution of each component, establishing DSFT as a robust and effective framework. Our work makes a clear contribution: by shifting from a uniform to a principled fine-tuning strategy, we can successfully inspire a model's deep understanding of highly structured domains.

\section{Method}
\label{sec:method}

\subsection{Background: Masked Diffusion Language Models}
\label{ssec:subhead}

Diffusion Large language models, a significant departure from auto regressive architectures \cite{li2022diffusion}, generate text through an iterative denoising process. The core idea involves a forward process that gradually corrupts a clean text sequence $\mathbf{x}_0 = (x_1, \ldots, x_L)$ by masking tokens, and a reverse process where a model learns to reconstruct the original sequence from its corrupted version.

During training, a noise level $t \sim \mathcal{U}[\epsilon, 1-\epsilon]$ is sampled, which determines the proportion of tokens to be masked. This creates a noisy input $\mathbf{x}_t$, where a subset of tokens is replaced by a special `[MASK]` token. The model $f_\theta$ is then trained to predict the original tokens at the masked positions, optimizing a cross-entropy loss:

\vspace{-1.5em}
\begin{equation}
\mathcal{L} = \mathbb{E}_{t,\mathbf{x}_0,\mathbf{x}_t}\left[\sum_{i=1}^L \mathbf{1}[\mathbf{x}_t^i = \text{[MASK]}] \cdot \text{CE}(f_\theta(\mathbf{x}_t)_i, x_{0,i})\right]
\end{equation}

This bidirectional training objective allows the model to use the full context, both preceding and succeeding, for reconstruction. While effective for general language, this standard approach treats all tokens uniformly, a suboptimal strategy for specialized domains like mathematics where certain tokens carry disproportionate semantic weight.

\subsection{DSFT: Guiding Mathematical Comprehension}
\label{sec:subhead}

DSFT is not an architectural change but a set of principles designed to address the shortcomings of uniform SFT. The methodology is designed to guide the model towards the unique "grammar" of mathematical text, helping it to recognize that numbers are not just tokens and that expressions have a sequential, logical structure. This is achieved by introducing non-uniformity into the masking process and the loss function to prioritize important information.

\subsubsection{Number-First Masking}
\label{sssec:subsubhead}

Mathematical understanding depends on numerical precision. To make the model more sensitive to the importance of numbers, we introduce a targeted masking strategy. In addition to the baseline uniform masking applied to the entire sequence, our method first identifies all tokens that represent numerical values. A fixed proportion of these identified numerical tokens are then additionally masked. This strategy significantly increases the frequency with which the model must reconstruct numbers from their surrounding context, compelling it to develop a more robust understanding of numerical relationships and computations.

\subsubsection{Span Masking}
\label{sssec:subsubhead}

Logical steps and mathematical expressions often form coherent, multi-token phrases like "x equals 3.14". To encourage the model to learn these local dependencies, we employ span masking. With a small probability, we randomly select a starting position in the sequence and mask a short, contiguous span of tokens. This forces the model to reconstruct entire phrases from the broader context, improving the coherence and logical flow of its generated text.

\subsubsection{Curriculum Masking}
\label{sssec:subsubhead}

To ensure training stability and guide the model from simple to more complex reconstruction tasks, we implement a curriculum learning strategy through an adaptive masking ratio. At the beginning of training, the overall proportion of masked tokens is low, presenting the model with a relatively easy denoising task. As training progresses, this ratio is linearly increased towards a predefined maximum. This gradual increase in difficulty allows the model to build foundational knowledge on simpler tasks before tackling more challenging reconstructions with sparser context, preventing early-stage training instability and promoting a more effective learning trajectory.

\subsubsection{Number Weighted Loss}
\label{sssec:subsubhead}

To further strengthen the learning signal from numerical content, we adjust the training objective itself. We apply a weighted cross-entropy loss that assigns greater importance to errors made on numerical tokens. Specifically, the loss contribution from each masked token is weighted based on whether it is numerical.

\begin{equation}
\mathcal{L}_{w} = \frac{\sum_{i \in \mathcal{M}} w_i \cdot \text{CE}(f_\theta(\mathbf{x}_t)_i, x_{0,i})}{\sum_{i \in \mathcal{M}} w_i}
\end{equation}

where $\mathcal{L}_{w}$ represents the weighted loss, $\mathcal{M}$ is the set of masked positions, and the weight $w_i$ is set to a value greater than 1 (e.g., 2.0) when the token $x_{0,i}$ is numerical, and 1.0 otherwise. This ensures that the model is more heavily penalized for inaccuracies in numerical prediction, directly prioritizing mathematical correctness during gradient updates.

\subsection{Why Do Numbers Matter? An Information-Theoretic View}
\label{sec:subhead}

The core principle of DSFT is to guide the model's attention towards semantically significant tokens. From an information-theoretic perspective, the "importance" of a token can be quantified by its information content, or entropy. The entropy $H(X)$ of a random variable $X$ with distribution $p(x)$ is given by:
\begin{equation}
H(X) = -\sum_{x \in \mathcal{X}} p(x) \log p(x)
\end{equation}
In a general English corpus, common words like "the" have a very high probability, and thus low information content. For example, if $p(\text{"the"}) \approx 0.05$, its contribution to entropy is small. In contrast, within the specialized corpus of mathematics, numerical tokens and operators form a much larger and more uniform distribution. The probability of any specific number is far lower than that of a common stopword, which means that its information content is significantly higher. A uniform fine-tuning objective, $\mathcal{L}_{SFT}$, implicitly assumes a uniform information distribution across all tokens, which is an inefficient allocation of the model's learning capacity.

The DSFT framework directly addresses this by creating an information-weighted objective. Our masking and loss-weighting strategies can be unified under a single objective that prioritizes high-entropy tokens. Let $p_{DSFT}(\mathcal{M}|\mathbf{x}_0)$ be our non-uniform masking distribution and $w_i$ be our loss weights. The DSFT loss is the following.
\begin{equation}
\mathcal{L} = \mathbb{E}_{\mathbf{x}_0, \mathcal{M} \sim p(\cdot|\mathbf{x}_0)} \left[ \sum_{i \in \mathcal{M}} w_i \cdot \text{CE}(f_\theta(\mathbf{x}_{\mathcal{M}})_i, x_{0,i}) \right]
\end{equation}
This objective function acts as a more efficient gradient steering mechanism. It focuses the model's learning capacity on the tokens that are hardest to predict and most critical for understanding the domain's underlying structure, which promotes a deeper and more robust understanding of mathematical patterns.

\subsection{Integrated Training Process}
\label{sec:subhead}

The four DSFT components are applied sequentially within a single training step. For each sequence in a batch, we first apply a base level of random masking. Then, the number-first, span, and curriculum masking strategies are applied additively, progressively increasing the set of masked tokens. To preserve the problem context, we ensure that tokens belonging to the input prompt are never masked. Finally, to maintain stable gradient flow in distributed training settings, we guarantee that at least one token is masked in every sequence. Figure \ref{fig:methodology_overview} provides a concrete example, illustrating how each of the four DSFT techniques is applied together to a mathematical word problem.

\begin{table*}[t]
\centering
\begin{threeparttable}
\caption{Main results on LLaDA-1.5 across mathematics, logic, and general tasks.}  % ← 先写 caption
\label{tab:ablation_study}
\small
{\renewcommand{\arraystretch}{0.8}
\begin{tabular*}{\textwidth}
{@{\extracolsep{\fill}} l *{9}{>{\centering\arraybackslash}p{1.2cm}} @{}}
\toprule
\multirow{2}{*}{\textbf{Method}} &
\multicolumn{3}{c}{\textbf{Mathematics}} &
\multicolumn{3}{c}{\textbf{Logic}} &
\multicolumn{3}{c}{\textbf{General}} \\
\cmidrule(lr){2-4} \cmidrule(lr){5-7} \cmidrule(lr){8-10}
& GSM8K & MATH & GPQA & BBH & ARC-C & WinoGrande & TruthfulQA & HellaSwag & MMLU \\
\midrule
LLaDA-1.5 (Base) & 75.59 & 23.74 & 29.69 & 51.27 & 56.57 & 71.74 & 48.14 & 71.61 & 64.21 \\
LLaDA-1.5 (SFT) & 75.72 & 23.54 & 30.01 & 51.36 & 56.60 & 71.74 & 47.98 & 71.84 & 64.12 \\
LLaDA-1.5 (DSFT) & 79.37 & 26.14 & 31.50 & 51.99 & 57.74 & 72.30 & 48.04 & 71.77 & 64.19 \\
\midrule
$\Delta_{\text{SFT}}$ &
${\color{red}\uparrow}0.17\%$ &
${\color{green}\downarrow}0.84\%$ &
${\color{red}\uparrow}1.08\%$ &
${\color{red}\uparrow}0.18\%$ &
${\color{red}\uparrow}0.05\%$ &
$-$ &
${\color{green}\downarrow}0.33\%$ &
${\color{red}\uparrow}0.32\%$ &
${\color{green}\downarrow}0.14\%$ \\[2pt]
$\Delta_{\text{DSFT}}$ &
${\color{red}\uparrow}\textbf{5.00}\%$ &
${\color{red}\uparrow}\textbf{10.11}\%$ &
${\color{red}\uparrow}\textbf{6.10}\%$ &
${\color{red}\uparrow}1.40\%$ &
${\color{red}\uparrow}\textbf{2.07}\%$ &
${\color{red}\uparrow}0.78\%$ &
${\color{green}\downarrow}0.20\%$ &
${\color{red}\uparrow}0.22\%$ &
${\color{green}\downarrow}0.03\%$ \\
\bottomrule
\end{tabular*}}
\begin{tablenotes}[flushleft]
\small
\item $\Delta$ shows relative improvement over baseline.
\end{tablenotes}
\end{threeparttable}
\end{table*}

\section{Experimental Setup}
\label{sec:setup}

\subsection{Evaluation Philosophy}

The goal of our evaluation is to assess if DSFT can inspire a model's structural understanding, a skill different from factual recall. While base dLLMs may have encountered problems from our test benchmarks during their extensive pre-training, our hypothesis is that they processed this data with a semantically-blind, uniform objective. Therefore, performance gains on these benchmarks should be caused by an improved *understanding* of mathematical and logical patterns, inspired by our principled guidance, rather than exposure to new information.

\subsection{Training Data}

We build a targeted fine-tuning dataset by combining three publicly available resources: MathQA \cite{amini2019mathqa} for mathematical problems, CommonsenseQA \cite{talmor2019commonsenseqa} for general question answering, and OpenBookQA \cite{mihaylov2018can} for science-based questions. This creates a diverse corpus totaling approximately 3.7 million tokens. This small data footprint is intentional; it emphasizes our hypothesis that developing deep structural understanding does not require massive new datasets, but rather a focused and principled guidance strategy. The diversity of the data encourages the model to develop a robust and generalizable understanding, rather than overfitting to a single task format.

\subsection{Evaluation Benchmarks and Configuration}

To measure the impact of our methodology, we evaluate the fine-tuned models on a comprehensive suite of nine benchmarks across three domains: Mathematics (GSM8K \cite{cobbe2021training}, MATH \cite{hendrycks2020measuring}, GPQA \cite{rein2024gpqa}), Logic (BBH \cite{suzgun2022challenging}, ARC-C \cite{clark2018think}, WinoGrande \cite{sakaguchi2021winogrande}), and General tasks (TruthfulQA \cite{lin2021truthfulqa}, HellaSwag \cite{zellers2019hellaswag}, MMLU \cite{hendrycks2021measuring}).

We apply our fine-tuning to the LLaDA-1.5 \cite{zhu2025llada} and Dream-7B \cite{ye2025dream} models. We use the following hyperparameters: learning rate $1 \times 10^{-6}$, batch size 64, sequence length 1024, and 1 epoch. All four DSFT techniques are enabled with a numerical loss weight of 2.0, a span length of 3, a span probability of 0.1, and a curriculum masking ratio increasing from 10\% to 20\%.

\section{Results and Analysis}
\label{sec:results}

\subsection{Main Results on LLaDA-1.5}
\label{sec:subhead}

As shown in Table \ref{tab:ablation_study}, our DSFT methodology delivers significant performance gains when applied to the LLaDA-1.5 model. On the mathematical benchmarks, DSFT achieves a substantial relative improvement of \textbf{5.00\%} on GSM8K and a remarkable \textbf{10.11\%} on MATH over the base model. This is not just a score increase; it indicates that the model is beginning to better understand the specific patterns of mathematical text, a direct result of the structured guidance provided by our fine-tuning approach. We also see a significant \textbf{6.10\%} improvement on GPQA.

This effectiveness extends to logical tasks, with a strong relative improvement of \textbf{2.07\%} on ARC-C. The consistent gains support our hypothesis that a simple, targeted fine-tuning strategy can effectively guide dLLMs to overcome their inherent non-sequential nature in structured domains. Standard Supervised Fine-Tuning (Base + SFT), in contrast, offers negligible or even negative effects, highlighting that a naive approach fails to develop the necessary understanding.

\begin{figure}[h]
    \centering
    \includegraphics[width=0.4\textwidth]{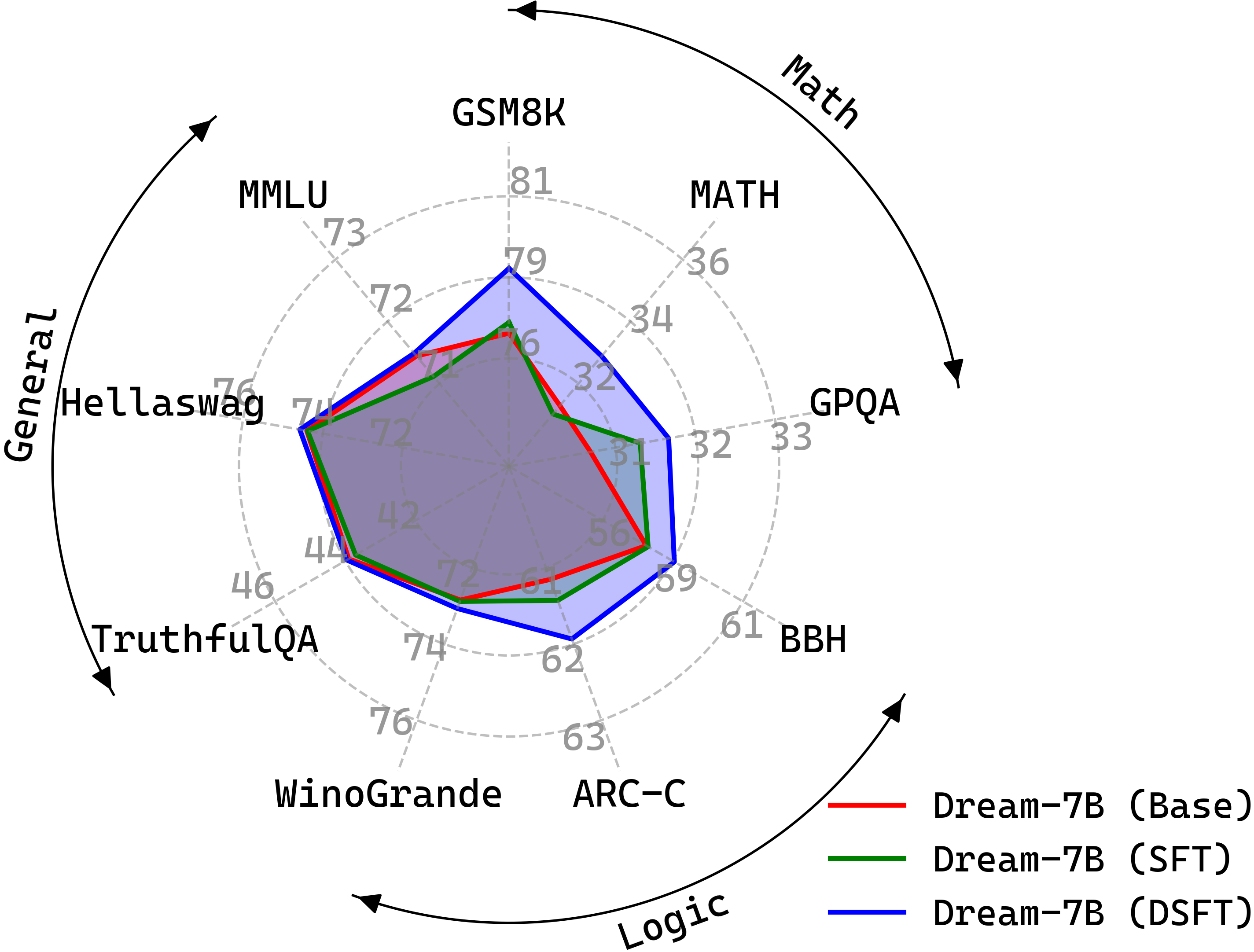}
    \caption{Performance comparison on Dream-7B }
    \label{fig:radar_plot}
\end{figure}

\subsection{Generalization Across Model Architectures}
\label{sec:subhead}

To demonstrate the architectural generalization of our methodology, we applied DSFT to Dream-7B, a dLLM from a different model family. The results, visualized in Figure \ref{fig:radar_plot}, show that DSFT provides a consistent performance improvement across a diverse set of nine benchmarks. The blue line, representing the DSFT-enhanced model, forms a consistently larger area than the baseline models, indicating an overall improvement. This successful application to a different model architecture strongly supports our claim that DSFT is not a model-specific set of tricks, but a general and effective methodology for fine-tuning dLLMs.

\subsection{Ablation Study: Deconstructing the Guidance Process}
\label{sec:subhead}

To understand how each component contributes to guiding the model, we conducted a systematic ablation study on LLaDA-1.5 (Table \ref{tab:ablation}). The baseline SFT shows very small improvement, confirming that undirected fine-tuning is insufficient for developing pattern understanding. 

\vspace{-1.5em}
\begin{table}[htb]
\centering
\caption{Ablation experiments LLaDA-1.5.}
\label{tab:ablation}
\small
{\renewcommand{\arraystretch}{0.8}
\begin{tabular}{@{}l|cc@{}}
\toprule
Component & GSM8K & ARC-C \\
\midrule
LLaDA-1.5 (Base) & 75.59 & 56.57 \\
LLaDA-1.5 (SFT) & 75.72 \ (${\color{red}\uparrow}0.17\%$) & 56.60 \ (${\color{red}\uparrow}0.05\%$) \\
+ Number masking &  76.21 \ (${\color{red}\uparrow}0.82\%$) & 56.85 \ (${\color{red}\uparrow}0.49\%$) \\
+ Span masking & 76.75 \ (${\color{red}\uparrow}1.54\%$) & 57.02 \ (${\color{red}\uparrow}0.80\%$) \\
+ Curriculum masking & 76.20 \ (${\color{red}\uparrow}0.81\%$) & 56.46 \ (${\color{green}\downarrow}0.19\%$) \\
+ Loss weighting & 77.18 \ (${\color{red}\uparrow}2.10\%$) & 56.52 \ (${\color{green}\downarrow}0.09\%$) \\
\midrule
LLaDA-1.5 (DSFT) & \textbf{79.37} \ (${\color{red}\uparrow}\textbf{5.00}\%$) & \textbf{57.74} \ (${\color{red}\uparrow}\textbf{2.07}\%$) \\
\bottomrule
\end{tabular}}
\end{table}

Introducing the DSFT components one by one reveals their individual impact on GSM8K and ARC-C. The two number-focused techniques, \textbf{Number-First masking} and \textbf{Numerical loss weighting}, provide the most significant individual improvements on GSM8K, with relative improvements of \textbf{0.82\%} and \textbf{2.10\%} respectively. This confirms that guiding the model to prioritize numbers is a critical first step. Interestingly, while some individual techniques like curriculum masking show a slight decrease on ARC-C, their inclusion in the complete method is critical for achieving the best overall performance. The full DSFT methodology, combining all four basic techniques, is clearly more effective when combined, creating a comprehensive "curriculum" that outperforms any partial one, achieving the final relative improvements of \textbf{5.00\%} on GSM8K and \textbf{2.07\%} on ARC-C.

% \subsection{Computational Efficiency}
% \label{sec:subhead}

% A key advantage of the DSFT methodology is its computational efficiency. Because our approach is based on simple modifications to the fine-tuning process rather than complex architectural changes, it introduces negligible overhead compared to standard SFT. The masking and loss weighting calculations are lightweight and fully parallelizable on modern hardware. Our entire fine-tuning process completes in just one epoch. This makes DSFT a practical and resource-efficient method for instilling deep mathematical understanding in any pre-trained dLLM.

\section{Conclusion and Discussion}
\label{sec:conclusion}

In this work, we addressed a core challenge in specializing large language models: the semantically-blind nature of standard fine-tuning. We proposed DSFT, a methodology that replaces uniform SFT with a set of principled guidance techniques to inspire a deeper understanding of mathematical and logical patterns. Our approach is supported by the information-theoretic insight that not all tokens are created equal in specialized domains. By guiding the model's attention to high-information features like numbers and local structures, we can significantly improve the mathematical and logical understanding of advanced dLLMs. Our experiments on small datasets confirm that DSFT is an effective and generalizable methodology that improves specialized understanding without compromising general language capabilities. The principle of moving from uniform to guided fine-tuning is widely applicable, and we hope this work inspires future research into more advanced, information-aware guidance strategies for a variety of structured domains.

\vfill\pagebreak

\bibliographystyle{IEEEbib}
\bibliography{refs}

\end{document}